\documentclass[10pt,twocolumn,letterpaper]{article}

 \usepackage[pagenumbers]{cvpr} 
\usepackage{xspace}
\usepackage{subcaption}
\usepackage{graphicx}
\usepackage[autolanguage]{numprint}
\usepackage[utf8]{inputenc}
\usepackage{microtype}
\usepackage{soul}
\usepackage{float}
\usepackage{textcomp, gensymb}
\usepackage{amssymb}
\usepackage{booktabs}
\usepackage{array}
\usepackage{relsize}
\usepackage[export]{adjustbox}

\usepackage[dvipsnames]{xcolor}

\newcommand\blfootnote[1]{%
\begingroup 
\renewcommand\thefootnote{}\footnote{#1}%
\addtocounter{footnote}{-1}%
\endgroup 
}

\definecolor{cvprblue}{rgb}{0.21,0.49,0.74}
\usepackage[pagebackref,breaklinks,colorlinks,citecolor=cvprblue]{hyperref}

\begin{document}

\def\name{DragScene~}

\author{
\textbf{Chenghao Gu$^1$} \quad 
\textbf{Zhenzhe Li$^2$} \quad 
\textbf{Zhengqi Zhang$^3$} \quad
\textbf{Yunpeng Bai$^4$}  \quad
\textbf{Shuzhao Xie$^1$} \\ \quad
\textbf{Zhi Wang$^{\text{\textbf{\textcolor{red}{*}}}1}$} \quad
\vspace{0.2cm} \\
$ ^1$Shenzhen International Graduate School, Tsinghua University \\
$ ^2$College of Artificial Intelligence, Xi’an Jiaotong University \\
$ ^3$School of Software, Beihang University \\
$ ^4$Department of Computer Science, The University of Texas at Austin \quad
}

\title{DragScene: Interactive 3D Scene Editing with Single-view Drag Instructions}

\twocolumn[{
\maketitle
\begin{center}
    \centering
    \vspace{-0.3cm}
    \captionsetup{type=figure}
    \vspace{-0.2cm}
    \includegraphics[width=\textwidth]{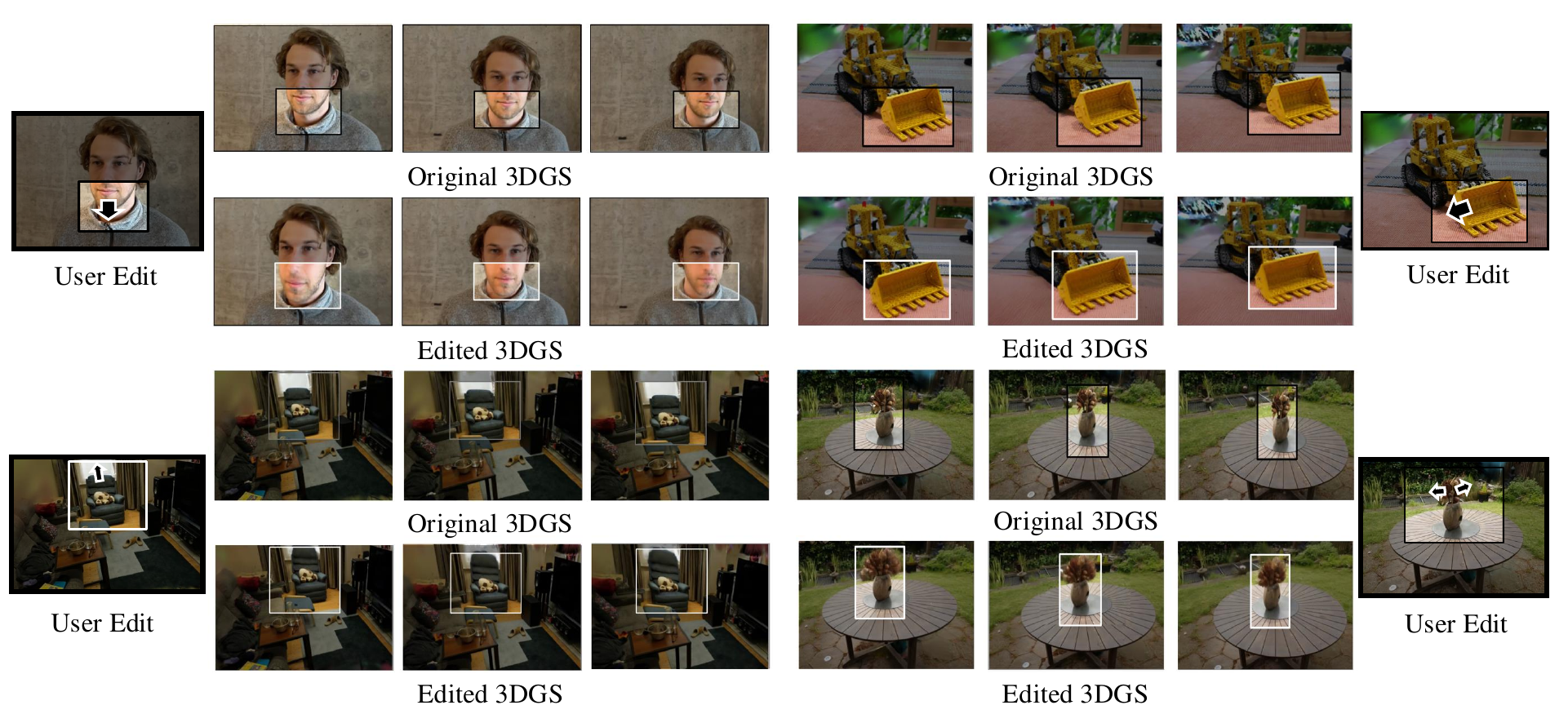}
    \captionof{figure}{\textbf{Results of DragScene.} DragScene successfully enables drag-based editing for 3D scenes. By following user-provided editing instructions (masks and points), our model seamlessly performs drag-style editing on the original 3D scene. 
    All the results presented above are based on 3D Gaussian Splatting (3DGS), demonstrating natural, creative, and view-consistent edits.
    }

    \label{fig:teaser}
\end{center}
}]

\blfootnote{\textcolor{red}{*} Corresponding author.}

\begin{abstract}
3D editing has shown remarkable capability in editing scenes based on various instructions. However, existing methods struggle with achieving intuitive, localized editing, such as selectively making flowers blossom. Drag-style editing has shown exceptional capability to edit images with direct manipulation instead of ambiguous text commands. Nevertheless, extending drag-based editing to 3D scenes presents substantial challenges due to multi-view inconsistency. To this end, we introduce DragScene, a framework that integrates drag-style editing with diverse 3D representations. First, latent optimization is performed on a reference view to generate 2D edits based on user instructions. Subsequently, coarse 3D clues are reconstructed from the reference view using a point-based representation to capture the geometric details of the edits. The latent representation of the edited view is then mapped to these 3D clues, guiding the latent optimization of other views. This process ensures that edits are propagated seamlessly across multiple views, maintaining multi-view consistency. Finally, the target 3D scene is reconstructed from the edited multi-view images. Extensive experiments demonstrate that DragScene facilitates precise and flexible drag-style editing of 3D scenes, supporting broad applicability across diverse 3D representations.
\end{abstract}    
\section{Introduction}
Recent advancements \cite{yuan2022nerf,kerbl20233d} in technologies for reconstructing 3D scenes from the real world have become increasingly important across a wide range of applications. To effectively meet user needs, it is essential that these reconstructed scenes support precise and controllable modifications. Consequently, 3D editing techniques have garnered significant attention, providing new opportunities for interactive user experiences and customized adjustments.

\begin{figure}[t]
    \centering
    \vspace{-.3cm}
    \includegraphics[width=0.44\textwidth]{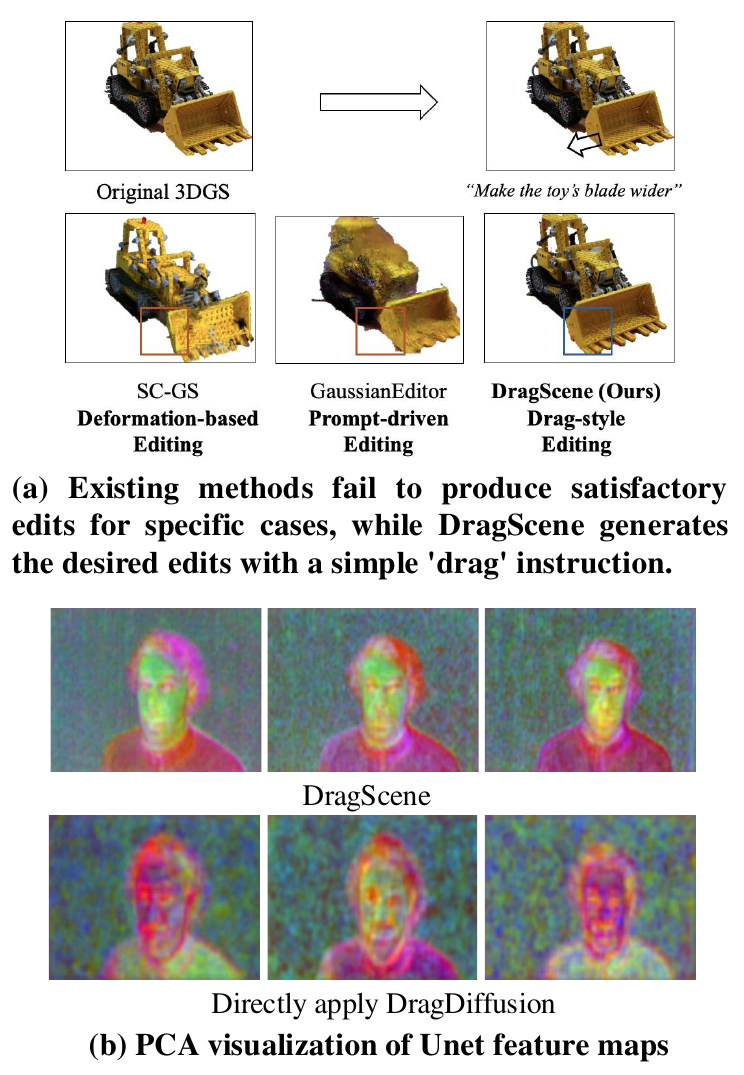}
    \vspace{-.3cm}
    \caption{\textbf{Our Motivation.} Comparison of DragScene and Directly applying DragDiffusion to multi-view images.  \textbf{(a)} illustrates that existing 3D editing methods fail to solve the specific editing task and DragScene performs well. \textbf{(b)} PCA visualization of Unet feature maps during the diffusion process. It demonstrates that directly applying 2D drag-style methods to multi-view images produces inconsistent features, whereas DragScene maintains multi-view feature consistency throughout the diffusion process.}
    \vspace{-.3cm}
    \label{fig:motivation_fig}
\end{figure}

Currently, many instruction-driven 3D editing methods offer excellent editing results, providing specific control over elements in a scene. Deformation-based editing methods \cite{xu2022deforming,jambon2023nerfshop} are able to deform 3D objects with techniques such as cages. However, they fail to achieve creative and precise editing effects due to the limitations of simple shape deformation and the lack of support for rich visual priors. As illustrated in Fig.~\ref{fig:motivation_fig}(a), we apply SC-GS \cite{huang2024sc} to perform the desired editing while producing unsatisfactory results.

Prompt-driven methods, based on pre-trained 2D diffusion models \cite{brooks2023instructpix2pix, rombach2022high}, can provide a wide variety of editing results based on text input, but are limited to overall appearance editing of the scene, lacking the ability of precise control over specific elements in the scene. The failure case conducted with GaussianEditor~\cite{wang2024gaussianeditor} in Fig.~\ref{fig:motivation_fig}(a) shows the limitation of prompt-driven methods.

Nowadays, a novel drag-style image editing method pioneered by DragGAN \cite{pan2023drag} has garnered widespread attention, where users are enabled to provide instructions like ``drag'' to edit a 2D image interactively and obtain realistic and rich editing results. However, it is challenging to propagate the 2D drag-style editing method a 3D scenes while maintaining multi-view consistency. As shown in Fig.~\ref{fig:motivation_fig}(b), we visualize the UNet feature maps \cite{ronneberger2015u} of multi-view images with principal component analysis (PCA) during the diffusion process when applying DragDiffusion~\cite{shi2024dragdiffusion} to multi-view images of 3D content. We observe that naively applying 2D drag-style editing to 3D scenes causes inconsistent features, leading to severe 3D discontinuities in editing.

To handle the aforementioned challenges, we introduce DragScene, a novel framework designed for flexible and controllable drag-based 3D editing while ensuring multi-view consistency. DragScene begins by rendering multi-view images of a 3D scene, and users are allowed to select a reference view to draw masks as editing regions and click handle points and target points to guide editing. Subsequently, we perform the desired editing on the reference view with 2D latent diffusion models, which optimize the latent of the original image to generate edits. To ensure consistent editing across multiple views, coarse point cloud representation of the edits is reconstructed as 3D clues to guide the further editing of other views. However, we observe that directly applying dense stereo methods to reconstruct the point cloud from a single edited image leads to significant deviations from the original scene. Therefore, we propose an effective optimization strategy for single-view point cloud reconstruction in the presence of scene edits, ensuring alignment with both the original scene and the edited region. To ensure consistency with the latent optimization process of the reference view, we assign the point cloud with the latent representation of the edited image. View-consistent latent maps can be rendered to further guide the latent optimization of other views, generating edited multi-view images with strong 3D consistency to reconstruct the final 3D representation of the edited scene. Extensive quantitative and qualitative experiments demonstrate that our method can achieve precise and creative editing results while maintaining excellent multi-view consistency. 

To summarize, our contributions can be summarized as follows:  1)  we propose an efficient 3D scene editing method named DragScene, which is the first drag-style 3D editing model applicable to real-world scenes.  2) we introduce a new mechanism to ensure multi-view consistency of 3D editing, effectively mapping 2D edits to multi-view images. 3) comprehensive experiments validate the versatility and generality of DragScene, which is highly competitive with existing editing methods.
\label{sec:intro}

\begin{figure*}
    \centering
    \includegraphics[width=0.98\textwidth]{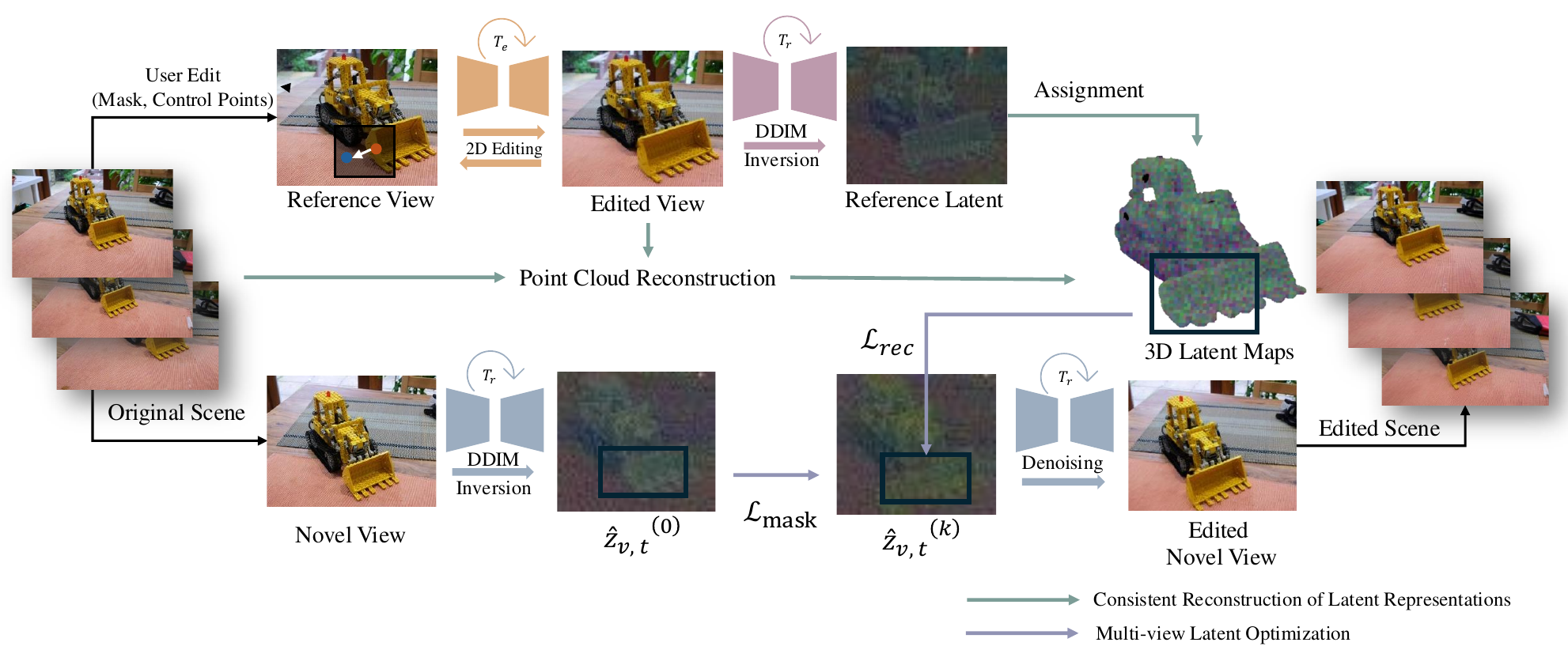}
    \vspace{-0.3cm}
    \caption{\textbf{Overview of DragScene.}
    Our approach consists of three steps: firstly, we apply a 2D drag-based diffusion model to edit the reference image and obtain the reference latent representation through DDIM inversion. Second, we perform consistent construction of the reference latent representation to obtain 3D latent maps. Finally, we apply the Inversion process to other views, further optimizing the images in latent space with the reconstructed 3D latent maps.
    }
    \vspace{-0.2cm}
    \label{fig:method_overview}
\end{figure*}

\section{Related Work}

\subsection{3D Scene Reconstruction}

Our framework is based on learnable 3D representations for scene reconstruction, which enables high-quality reconstruction through the learning of multi-view image information. The first neural network-based model for 3D reconstruction, Neural Radiance Field (NeRF)~\cite{mildenhall2021nerf}, brings groundbreaking advancements to 3D reconstruction. NeRF and its variants~\cite{chen2021mvsnerf,barron2022mip,fridovich2022plenoxels,guo2022nerfren,liu2020neural,yu2021plenoctrees, muller2022instant} enable accurate learning accurate scene representation through multi-view images. 3D Gaussian Splatting (3DGS)~\cite{kerbl20233d} proposed in 2023 surpasses NeRF in both reconstruction quality and speed, setting a new trend. Follow-up work~\cite{turkulainen2024dn,huang20242d,lyu20243dgsr} continues to introduce more effective methods for 3D scene reconstruction.

\subsection{3D Scene Editing}

3D editing methods allow users to modify the reconstructed scene according to specific editing requirements. Precise control over the appearance of a scene is achieved through two main methods: deformation-based methods and diffusion-based methods.

Deformation-based methods \cite{chen2023neuraleditor,liu2021editing,yang2022neumesh,yuan2022nerf, huang2024sc} leverage efficient techniques, such as cages \cite{peng2022cagenerf, xu2022deforming, jambon2023nerfshop}, to deform the 3D representation into target positions, enabling edits like movement and reshaping. However, due to the inherent limitations of simple deformation, these methods often lack creativity and struggle to modify fine details.

Diffusion-based methods allow for a direct description of editing requirements to change the overall appearance of the original scene. Current methods \cite{jambon2023nerfshop,haque2023instruct,chen2024consistdreamer,wang2024gaussianeditor,dong2024vica,wang2025view,chen2024dge,chen2024proedit} leverage pre-trained 2D diffusion models \cite{rombach2022high,brooks2023instructpix2pix} to guide the update of the 3D representation. While these methods allow for flexible and realistic edits, they often lack the precision needed to target specific details, which highlights the importance of our DragScene.

\subsection{Drag-style Editing}

Drag-style editing allows users to make precise adjustments to specific elements within an image. DragGAN \cite{pan2023drag} enhances a pre-trained GAN \cite{goodfellow2020gan} with feature-based motion supervision and point tracking for effective image editing.  DragDiffusion \cite{shi2024dragdiffusion} extends drag-style editing to the diffusion models and enables efficient spatial control by optimizing the latent during the diffusion process, while DragonDiffusion \cite{mou2023dragondiffusion} improves consistency using feature correspondence and visual cross-attention. SDE-Drag \cite{nie2023blessing} and RegionDrag \cite{lu2024regiondrag} employ copying and pasting methods within the diffusion latent space to achieve an efficient editing process. 

Meanwhile, drag-style editing can be extended beyond images. Drag-based video editing methods \cite{deng2025dragvideo,teng2023drag} effectively apply drag-style editing to videos, ensuring consistency across frames. MVDrag3D \cite{chen2024mvdrag3d} enables drag-style editing for 3D objects, offering a flexible and creative 3D editing approach. Despite these advancements, extending drag-style editing to real 3D scenes with complex backgrounds and details remains a challenge. Our DragScene addresses these challenges with the powerful generative capabilities of diffusion models and 3D clues derived from point-based representations.

\label{sec:r_w}
\section{Methodology}
\label{sec:method}
In this section, we formally present the DragScene method. Given a reconstructed real-world 3D scene, users provide masks and pairs of dragging points on the reference view to perform editing across the entire 3D scene. Sec.~\ref{sec:reference edit} introduces the drag-style editing on the reference view and the acquisition of the reference latent representation. Sec.~\ref{sec:cclr} outlines the consistent reconstruction of the latent representation, enabling editing operations that are tailored to multiple views. Sec.~\ref{sec:latent opt} discusses the specifics of latent optimization supervised by the reconstructed 3D latent maps for novel views. Finally, Sec.~\ref{sec:rec} details the reconstruction of the target 3D scene based on the edited view-consistent images.

\subsection{Reference View Editing}
\label{sec:reference edit}
Diffusion-based drag-style editing models focus on manipulating precise content editing for images, which is achieved by exerting controllable modifications to the latent representation during the diffusion process. In this paper, we employ DragDiffusion \cite{shi2024dragdiffusion} as the editing method on the reference view due to its suitable latent optimization strategy and powerful editing ability.

As a preliminary step, drag instructions and mask $M$ are provided by users according to the reference image $I_0$ rendered from the original 3D scene. A DDIM inversion process \cite{song2020denoising} is applied on the image and obtain a diffusion latent $\hat{z}_{t_{e}}^{(0)}$ at a certain step $t_{e}$. Through two steps namely Motion Supervision and Point Tracking, the diffusion latent is optimized through $m$ steps to $\hat{z}_{t_{e}}^{(m)}$, and $F ( \hat{z}_{v_0, t_{e}}^{(k)} )$ is the UNet output feature map. Also, two additional techniques, namely identity-preserving fine-tuning and reference-latent-control, are used to preserve the identity of the original scene. Following this, the image of the reference view is edited to the desired result $\hat{I}_0$.

We found that the latent representation obtained from the DragDiffusion process contains excessive noise, which makes it difficult to guide the subsequent optimization process. To derive a suitable representation in latent space, we perform an additional inversion process with a specific step $t_{r}$ on the edited image to obtain the latent representation $\hat{z}_{t_{r}}$.

\begin{figure}
    \centering
    \includegraphics[width=0.47\textwidth]{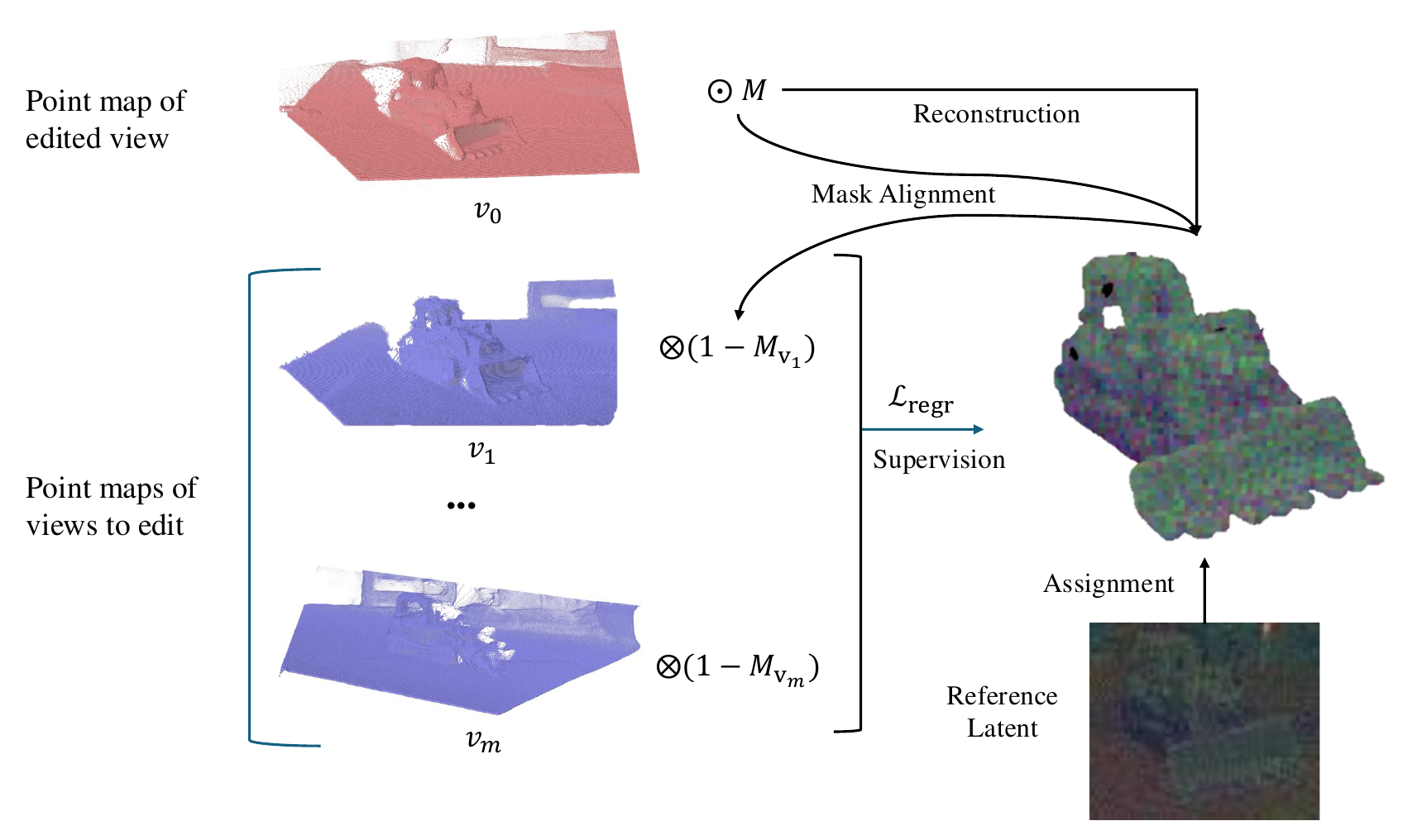}
    \vspace{-0.4cm}
    \caption{\textbf{Consistent Reconstruction of Latent Representations.} To facilitate consistent multi-view latent optimization, we apply DUSt3R to reconstruct the coarse 3D point cloud, with aligned masks assisting in the optimization process. The latent representation of the reference image is assigned to the point cloud to obtain the 3D latent maps.}
    \vspace{-0.1cm}
    \label{fig:method_2}
\end{figure}

\subsection{\normalsize \textbf{Consistent Reconstruction of Latent Representations}}
\label{sec:cclr}
Before editing other views of the original scene, we perform a consistent reconstruction of latent representation based on the prior diffusion process. This module, as detailed in Fig.~\ref{fig:method_2}, ensures that the edits are propagated consistently across different views within the scene.

Initially, DUSt3R \cite{wang2024dust3r} serves as the basic model to estimate the geometry information of the edits. To ensure that the coarse 3D clues of the edits do not deviate significantly from the original scene, we select the edited reference image $I_0$ and a sparse set of original images from other views, $\mathcal{I} = \{I_1, I_2, \dots, I_s\}$, as the input.

During the point cloud reconstruction process, we obtain a set of pointmaps $\{X_0, X_1, \ldots, X_s\}$ based on the edited image $I_0$ and the image set $\mathcal{I}$. $X_{n,m}$ represents the pointmap $X_n$ from view $v_n$, expressed in the coordinate frame of view $v_m$. To ensure consistency in the edited region, the mask $M$ provided by the user in the reference view $v_0$ is also transformed to view $v_m$. The transformation is as follows:

\begin{equation}
X_{n,m} = P_{m}{P_{n}}^{-1} h(X^n),
\end{equation}

\begin{equation}
M_{v_m} = P_{m}{P_{0}}^{-1} h(M),
\end{equation}
where \( P_m \) and \( P_n \) are the world-to-camera poses for images \( n \) and \( m \) respectively. The function \( h \) is a homogeneous mapping defined as \( h: (x, y, z) \rightarrow (x, y, z, 1) \).

It needs to be ensured that the unmasked regions align with all views, while the masked regions align with the edits of the reference view. The regression loss for pixel $i$ on view $m$ is defined as follows:

\begin{equation}
\begin{aligned}
\mathcal{L}_{\text{regr}}(m,i) = 
&\frac{1}{s+1}\sum_{n=0}^s\| (\frac{1}{z}X_i^{m} - \frac{1}{z}\bar{X}_i^{n,m}) \odot (\mathbf{1}
 - M_{v_m}) \|  \\
&+ \| (\frac{1}{z}X_i^{m} - \frac{1}{z}\bar{X}_i^{0,m}) \odot M_{v_m} \| ,
\end{aligned}
\end{equation}
where $\odot$ denotes the Hadamard product.

After reconstructing the coarse point cloud $\mathcal{P}$, we assign the latent representation $\hat{z}_{t_{r}}$ and mask $M$ to the point cloud to obtain the attributed point cloud $P_\mathcal{Z}$ as the 3D latent representation and $P_\mathcal{M}$ as the 3D mask. For the set of views $\mathcal{V} = \{v_1, v_2, \ldots, v_s\}$ designated for editing, we can obtain consistent latent maps and editing masks through rendering, which serves to guide the subsequent latent optimization of other views.

\subsection{Multi-View Latent Optimization}
\label{sec:latent opt}

This section provides a detailed description of the multi-view latent optimization process in DragScene, which optimizes the image latent during the diffusion process with consistent latent representations and masks.

Inspired by the latent optimization in DragDiffusion \cite{shi2024dragdiffusion}, we first perform a DDIM inversion on the novel view image $I_{v_i}$ to obtain the diffusion latent representation \( z_{v_i, t_{r}} \) at step $t_{r}$. We denote the initial latent representation as \( z_{v_i, t_{r}}^{(0)} \). The latent map \( R_{v_i}(\mathcal{P_\mathcal{Z}}) \) and the editing mask \( R_{v_i}(\mathcal{P_\mathcal{M}}) \) are obtained through the rendering process \( R_{v_i}(\cdot) \) that projects the values of point cloud to image given view $v_i$. 

The loss function at the \( k \)-th iteration is:
\begin{equation}
\mathcal{L}_{\text{total}} = \mathcal{L}_{\text{rec}} + \lambda \mathcal{L}_{\text{mask}},
\end{equation}
where
\begin{equation}
\mathcal{L}_{\text{rec}} = \left\|(z_{v_i, t_{r}}^{(0)} - \mathcal{S}_g(R_{v_i}(\mathcal{P_\mathcal{Z}})))\odot R_{v_i}(\mathcal{P_\mathcal{M}}) \right\|_1,
\end{equation}
and
{
\footnotesize 
\begin{equation}
\mathcal{L}_{\text{mask}} = \left\| 
\left( \hat{z}_{v_i, t_{\text{r}}-1}^{(k)} - \mathcal{S}_g\left( \hat{z}_{v_i, t_{\text{r}}-1}^{(0)} \right) \right)
\odot 
\left(\mathbf{1} - R_{v_i}(\mathcal{P}_\mathcal{M}) \right)
\right\|_1.
\end{equation}
}

Here, \( \hat{z}_{v_i, t_{r}}^{(k)} \) represents the latent at the \( k \)-th iteration, and \( \mathcal{S}_g(\cdot) \) denotes the stop-gradient operator. In each iteration, \( \hat{z}_{v_i, t_{r}}^{(k)} \) is optimized by performing a single gradient descent step to minimize the total loss function \( \mathcal{L}_{total} \):

\begin{equation}
    \hat{z}_{v_i, t_{r}}^{(k+1)} = \hat{z}_{v_i, t_{r}}^{(k)} - \sigma \cdot \frac{\partial \mathcal{L}_{\text{total}} (\hat{z}_{v_i, t_{r}}^{(k)})}{\partial \hat{z}_{v_i, t_{r}}^{(k)}},
\end{equation}
where \( \sigma \) is the learning rate for latent optimization. During the optimization, \( \mathcal{L}_{\text{rec}} \) aligns the latent representation with the target latent map, while \( \mathcal{L}_{\text{mask}} \) constrains the unmasked regions to stay close to the original latent representation.

Then, we conduct the diffusion process \( f_{t_{r} \to 0} \) with the optimized latent representation \( \hat{z}_{v_i, t_{r}}^{(m)} \) after $m$ iterations to obtain \( \hat{z}_{v_i, 0} \). Then the edited image \( \hat{I}_{v_i} \) can be generated through decoding \( D(\hat{z}_{v_i, 0}) \) .

\subsection{Reconstruction of Edited Scenes}
\label{sec:rec}

After finishing the multi-view consistency optimization, we employ these 2D views to edit the corresponding 3D scene. Here, we take 3D Gaussian Splatting (3DGS) \cite{kerbl20233d} as an example due to its exceptional performance and speed in reconstructing unbounded scenes, and it has been widely adopted in a variety of applications~\cite{charatan23pixelsplat,zhou2024drivinggaussian,lu2024manigaussian,xie2024mesongs}. For the original 3DGS scene, we define a set of camera poses to render a set of multi-view images of the scene. Then, the steps introduced above are conducted to generate consistent edited multi-view images, which are then used for further 3DGS reconstruction. DragScene successfully propagates the drag-style edits performed on the reference view to the whole 3D representation, generating the target 3D scene that adheres to the editing instructions.

DragScene is a general 3D scene editing method, not tied to one specific 3D representation. It achieves 3D-consistent editing solely based on image information, which gives significant potential for extensibility to newly emerged 3D reconstruction models.

\section{Experiment}

\begin{figure*}
    \centering
    \includegraphics[width=\textwidth]{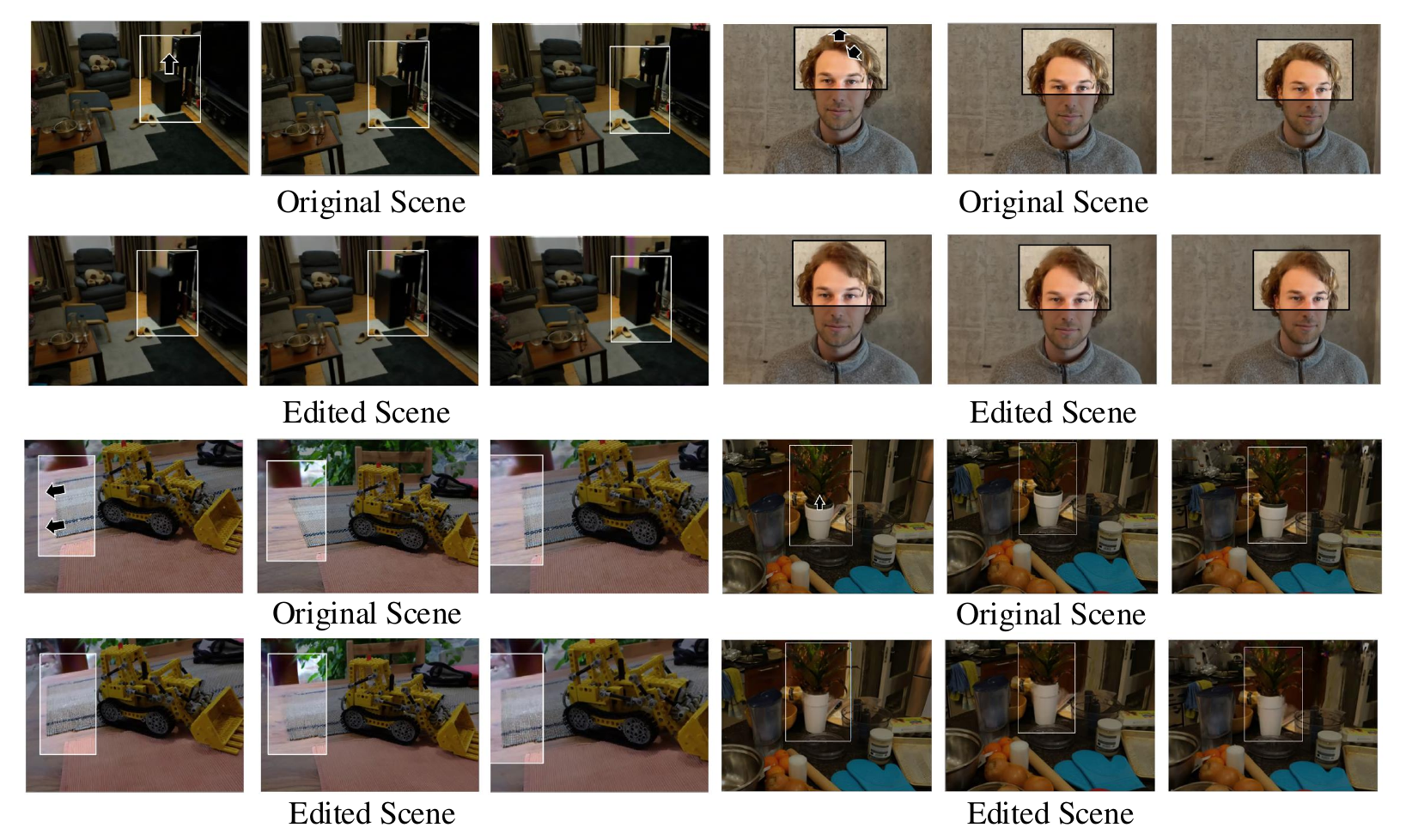}
    \vspace{-.4cm}
    \caption{\textbf{More results of DragScene.} We present various views of both the original and edited scenes. All scenes are reconstructed using 3D Gaussian splatting.
    }
    \vspace{-.3cm}
    \label{fig:results_all}
\end{figure*}

\begin{figure*}
    \centering
    \includegraphics[width=0.92\textwidth]{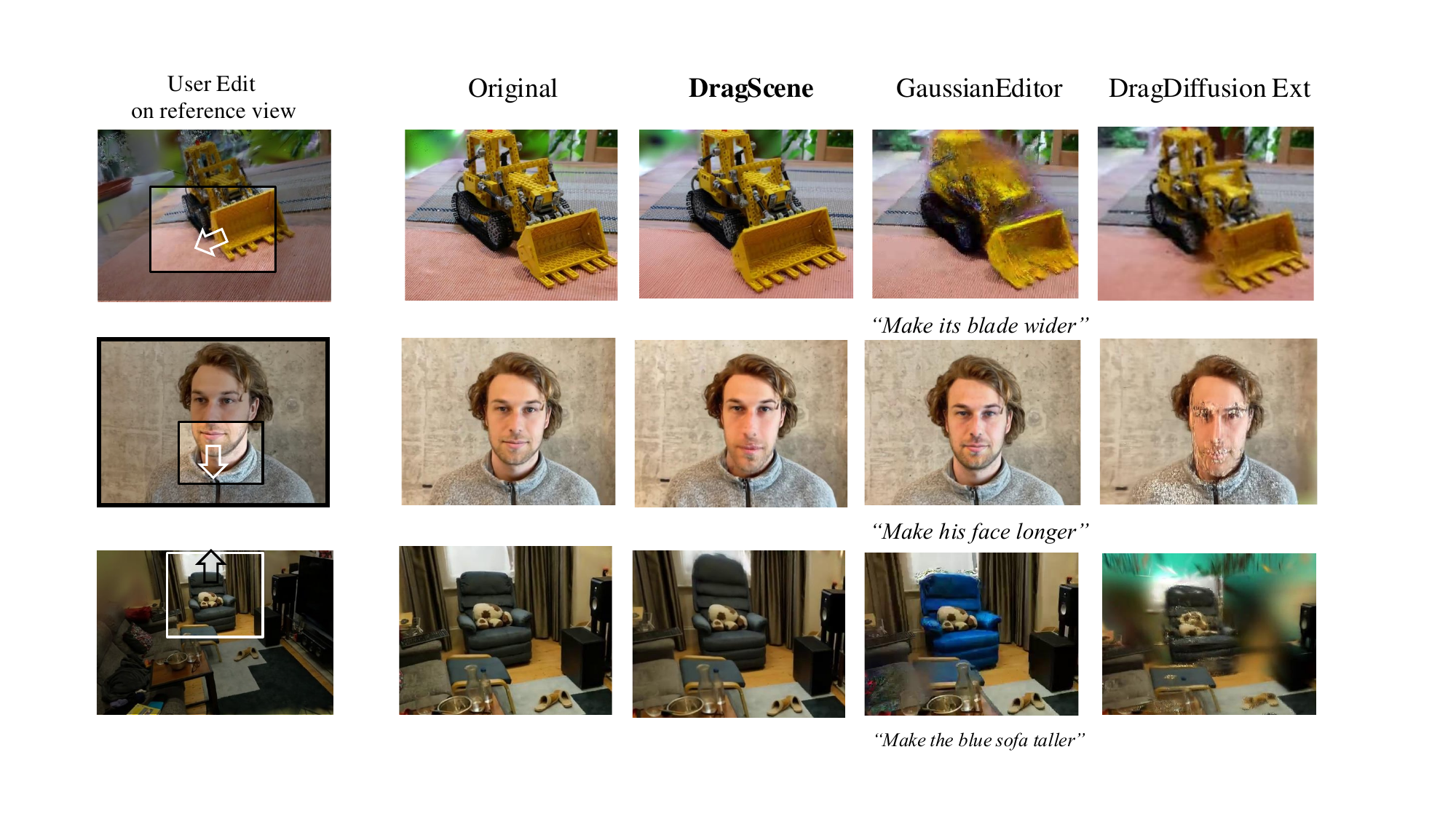}
    \vspace{-.3cm}
    \caption{\textbf{Comparisons between DragScene and other methods.} We show the rendering view of scenes edited with different models. For fairness, we provide different forms of editing instructions with the same editing intent.
    }
    \vspace{-.2cm}
    \label{fig:results_compare}
\end{figure*}

\subsection{Implementation Details}
To build our diffusion model, we adopt Stable Diffusion v1.5 \cite{rombach2022high} as the base model for the editing operation in DragScene. To enhance the multi-view consistency, we adopt the identity-preserving fine-tuning and reference latent control in DragDiffusion. We set the LoRA\cite{hu2021lora} rank to 16 and inject LoRA into the projection matrices of the query, key, and value tensors of every attention module. We then fine-tune LoRA with 80 iterations, using the AdamW optimizer\cite{diederik2014adam} with a learning rate of $5 \times 10^{-4}$. During DDIM Inversion, we set the total number of steps $t_{total}$ to 50, with $t_e=35$ and $t_r=20$.

We base the implementation of 3DGS on Threestudio\cite{threestudio2023} and employ the optimized renderer from \cite{kerbl20233d} for Gaussian rendering.  For editing, we set a specific camera trajectory to capture the original image set with about 20 images for editing and reconstructing the edited scene based on the preset camera poses. 

\begin{figure}
    \centering
    \includegraphics[width=0.47\textwidth]{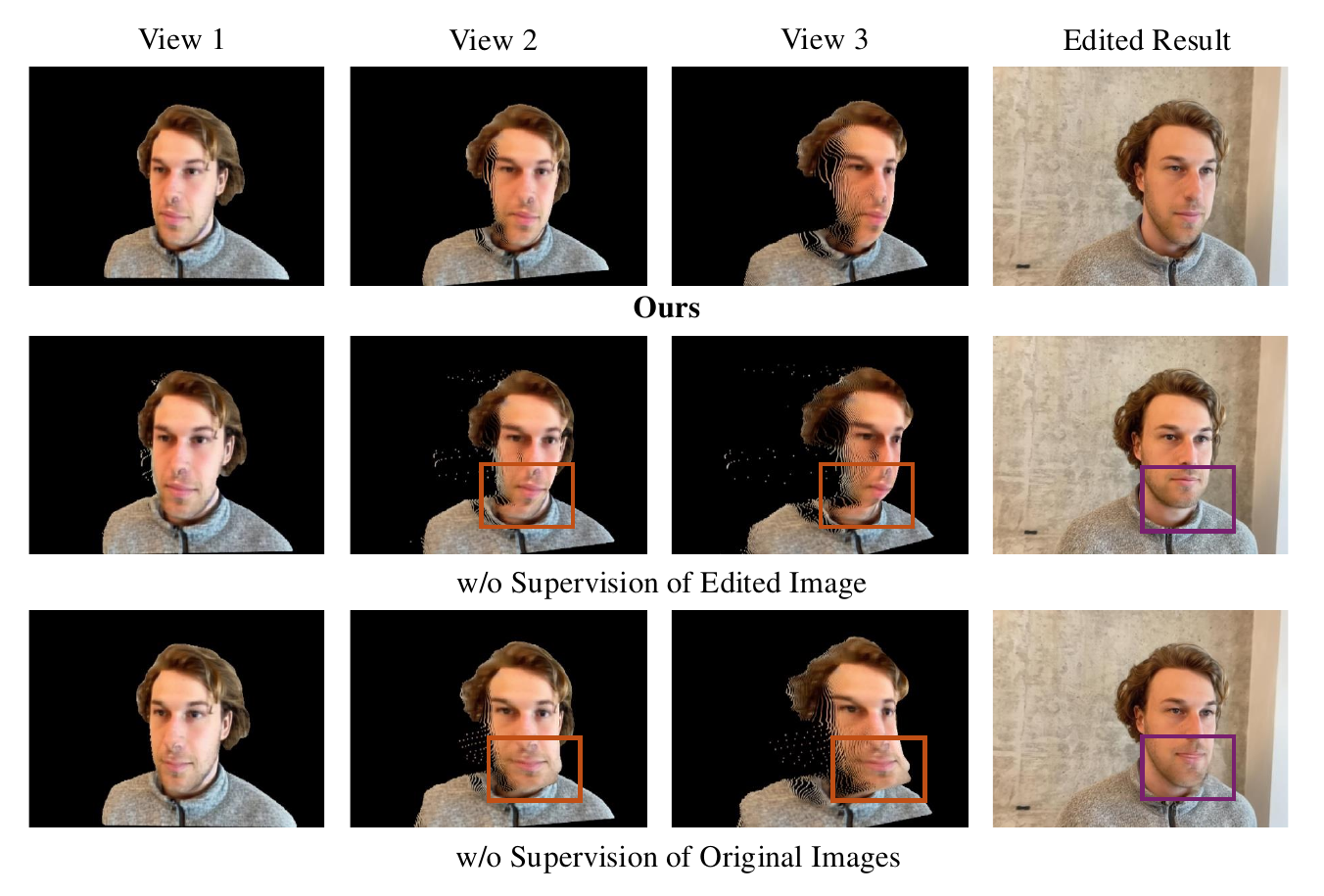}
    \vspace{-0.1cm}
    \caption{\textbf{Ablation on reconstruction of 3D clues.} The first three columns display the reconstructed point cloud of the reference view, and the fourth column shows the subsequent editing result. As shown, without the supervision of edited images, the predicted point cloud fails to accurately represent the target edited result, leading to mismatches in the latent features. Meanwhile, without the supervision of original images, the point cloud prediction deviates from the original scene. In contrast, DragScene's point cloud reconstruction successfully generates accurate coarse 3D clues.}
    \vspace{-0.1cm}
    \label{fig:ablation_3}
\end{figure}

\subsection{Experiment Setup}

\textbf{Datasets}
Recent advancements in drag-based editing techniques have highlighted a gap in available benchmark datasets, and several high-quality drag-style image editing datasets have recently emerged. However, a standardized dataset for evaluating drag-style edits in 3D scenes remains absent. To address this, we conduct our experiments on the Instruct-NeRF2NeRF \cite{haque2023instruct} Dataset and Mip-NeRF 360 Dataset \cite{barron2022mip} which are mostly used in real-world 3D editing with various 3D scenes. To comprehensively assess the effectiveness of our method, we perform diverse editing operations on various objects within 3D scenes containing faces, scenery, furniture, and more.

\textbf{Baseline}
As DragScene is the first method developed specifically for drag-style 3D scene editing, no universally recognized baselines exist for direct comparison. Therefore, to ensure an equitable comparison, we construct two comparative baselines. The first is GaussianEditor \cite{wang2024gaussianeditor}, a prompt-based 3D editing method, with carefully designed prompts to align with our targets. The second method applies DragDiffusion
\cite{shi2024dragdiffusion} directly on multi-view images, named DragDiffusion-3D extension. For fairness, we unify the 3D reconstruction for both DragScene and the baselines and the same drag instruction is applied.

\subsection{Qualitative Evaluation}
As shown in Fig.~\ref{fig:results_all}, our method surpasses other methods in both editing quality and controllability. Our framework demonstrates high-quality and precise editing capabilities in real-world 3D scenes, enabling users to achieve targeted editing effects with masks and control points provided from a single viewpoint. As we can see, the editing results are not limited to simple shape deformation and exhibit a broader range of complex generation. The edited scenes also exhibit excellent 3D consistency from multiple views and remain free from noticeable artifacts or failed reconstruction. Extensive examples highlight DragScene’s ability to effectively handle complex edits in 3D scenes, addressing limitations in other methods. Additional results can be found in the supplementary materials.

Meanwhile, we present a comparative analysis against other methods, as depicted in Fig.~\ref{fig:results_compare}. It can be observed that the direct application of the DragDiffusion model to multi-view images leads to 3D inconsistencies and artifacts. Furthermore, compared to prompt-based editing techniques like GaussianEditor \cite{wang2024gaussianeditor}, it is clear that GaussianEditor fails to deliver satisfactory editing results based on the provided instructions. Moreover, Dragdiffusion-3D Extension suffers from distortion and artifacts due to multi-view inconsistency.

\subsection{Ablation on Reconstruction of 3D Clues}
We conduct an ablation study on the reconstruction of 3D clues. During the optimization process of building a point cloud based on Dust3R, the edited reference image serves as the anchor view and the supervision of the original images is applied. The first row in Fig.~\ref{fig:ablation_3} demonstrates that the coarse point cloud reconstruction method used in DragScene effectively preserves the original appearance of the scene.

When directly employing the 3D clues of the original scene (such as the original point cloud of 3DGS) instead of the newly generated point cloud from the edited image, incorrect region correspondence occurs during the point cloud assignment, due to the absence of the newly updated point cloud regions. For the example shown in the second row of Fig.~\ref{fig:ablation_3}, the mouth area in the original point cloud shifts toward the neck region after a drag operation, resulting in feature mismatches. Additionally, without supervision from the multi-view images of the original scene, the point cloud reconstructed using only the reference image exhibits distortions in unseen regions as shown in the third row of Fig.~\ref{fig:ablation_3}, resulting in discrepancies with the original scene.

\begin{figure*}
    \centering
    \vspace{-.3cm}
    \includegraphics[width=\textwidth]{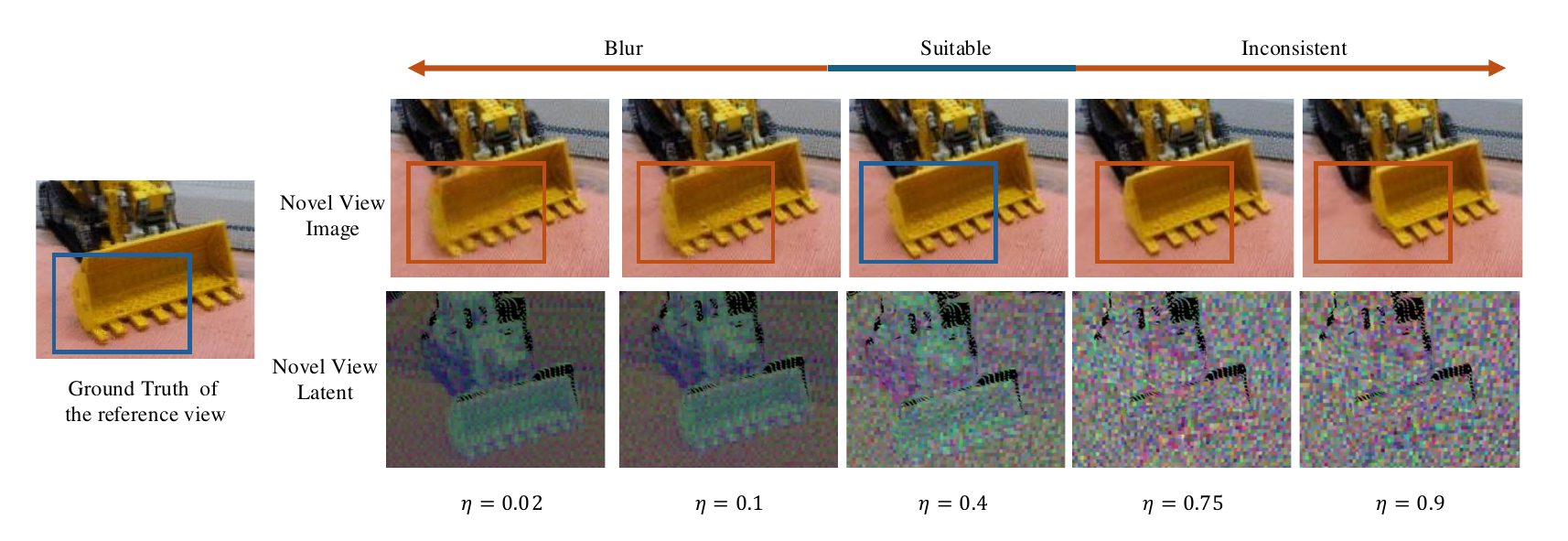}
    \caption{\textbf{Ablation Study on the the Inversion Strength} $\eta = t_{opt} / t_{total}$. Results of high quality and multi-view consistency are obtained when $\eta$ is around 0.4. When the value of $\eta$ is smaller, the noise in the latent representation decreases, leading to a more blurred output after optimization. On the other hand, when $\eta$ is larger, the noise in latent representation increases, causing a greater deviation from the target result and the emergence of artifacts. When $\eta$ is around 0.4, the quality of the optimized result is higher and closer to the target outcome.
    }
    \vspace{-.2cm}
    \label{fig:ablation_1}
\end{figure*}

\begin{figure}[!t]
    \centering
    \includegraphics[width=0.45\textwidth]{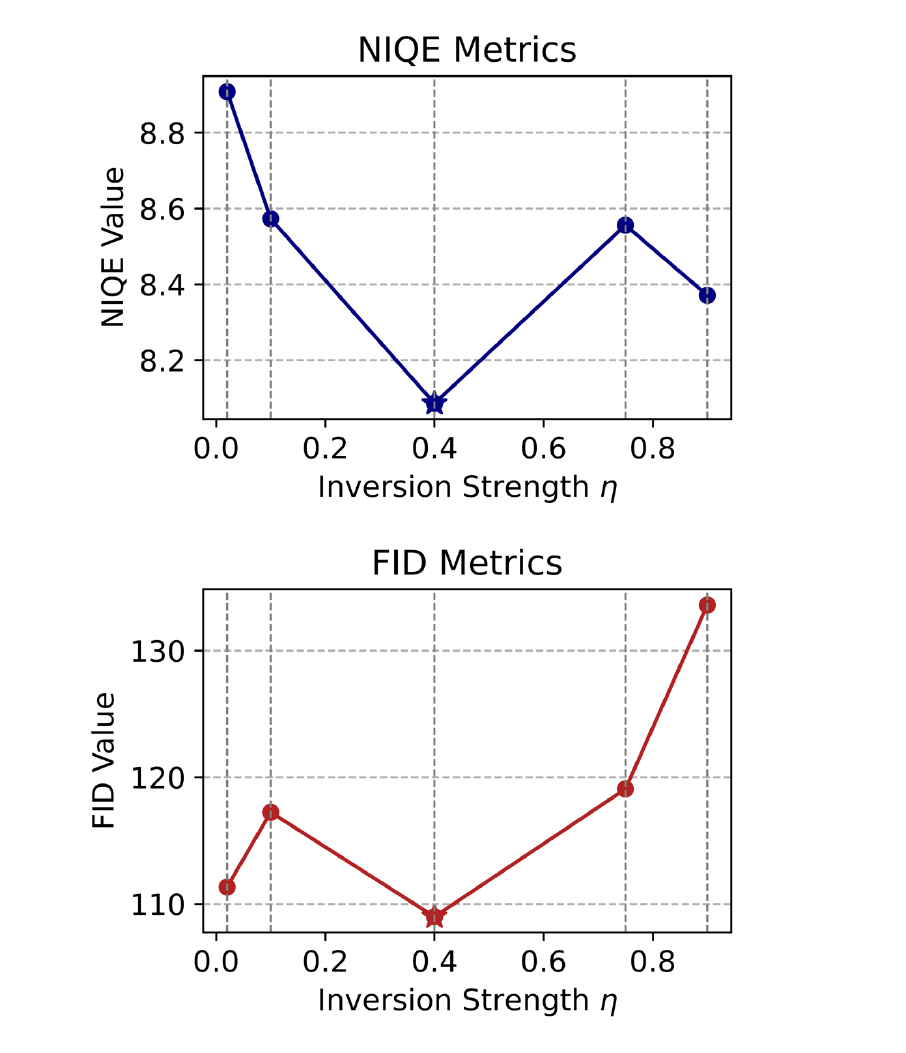}
    \caption{\textbf{Ablation study on the Inversion strength }$\eta$. NIQE Score ($\downarrow$) and FID Score ($\downarrow$) are reported. When the inversion strength \(\eta\) reaches 0.4, both metrics achieve their optimal performance.}

    \label{fig:ablation_2}
\end{figure}

\subsection{Ablation Study on Inversion Process}
We perform an ablation study to analyze the impact of the number of inversion steps $t_{r}$ to optimize the latent of multi-view images. We schedule the number of total DDIM steps $t_{total}$ to 50 and perform optimization at step $t_{r} = t_{total} * \eta$, where $\eta$ is the inversion strength that controls the inversion process ($\eta= 1$ indicates pure noise, and $\eta = 0$ represents the original image).

As shown in Fig.~\ref{fig:ablation_1}, when $\eta$ is relatively small, the noise in the latent representation is insufficient, resulting in lower image quality in the edited regions of the generated image. As $\eta$ increases, the latent representation approaches noise, resulting in improved image quality. However, when $\eta$ is too large, the result exhibits considerable deviation from the target and suffers from artifacts, which may cause inconsistencies across multiple views. It is evident that when $\eta$ is around 0.4, the generated image achieves optimal quality, with the content closely matching the editing target.

Subsequently, we evaluate the quality and consistency of the generated images from novel viewpoints by taking the edited view as a reference. Since there is no ground truth for evaluation, we use a no-reference image quality metric named NIQE Score \cite{mittal2012making} to assess the quality of the generated images. Additionally, we calculate the FID score \cite{heusel2017gans} between the generated images on the novel views and the edited reference view image to assess the degree of deviation from the editing target. As shown in Fig.~\ref{fig:ablation_3}, the optimal results are achieved when $\eta$ is around 0.4.

\section{Conclusion}
In our research, we introduce DragScene, a novel 3D editing method that harnesses the powerful generative capabilities of diffusion models and the coarse 3D clues provided by point-based representations. Our work propose a novel 3D editing paradigm that effectively overcomes the limitations of previous methods and resolves the challenge of multi-view inconsistency in 3D editing. Users are allowed to perform controllable edits on complex 3D scenes interactively through simple editing instructions. DragScene enables flexible and creative drag-style editing on diverse 3D scenes while maintaining high fidelity and multi-view consistency. We hope that DragScene will inspire exciting new applications and drive innovative research to develop more user-friendly methods for 3D editing. 

\textbf{Limitation.}
Despite the advantages of our method, there are still several limitations to consider. Firstly, the effectiveness of editing is limited to the performance of 2D diffusion models when performing drag-based editing on the reference image. For example, DragDiffusion \cite{shi2024dragdiffusion} may fail in some cases, such as opening the mouth of a man. Further research is encouraged to extend DragScene to support a broader range of 2D drag-based editing models. Meanwhile, the method may produce unsatisfactory results when the view range is too large, due to a lack of sufficient 3D clues. This issue is similar to that identified in other works such as \cite{yu2024viewcrafter} which also rely on coarse 3D clues to guide generation.

\textbf{Future Work.}
In the future, additional  research directions worth research. First, one promising direction for DragScene is to leverage large language models to use text prompt to guide drag instructions for 3D scene editing, enabling more user-friendly editing. Moreover, the editing method for static scenes can be further extended to dynamic scenes, allowing for controllable editing of more complex scenes. Furthermore, more editing paradigms can be explored based on the framework of our method, enabling more efficient editing on 3D scenes.
{
    \small
    \bibliographystyle{ieeenat_fullname}
    \bibliography{main}
}

\end{document}